\documentclass[11pt]{article}
\usepackage{acl}
\usepackage{times}
\usepackage{latexsym}
\usepackage{graphicx}
\usepackage{booktabs}
\usepackage{amsmath}
\usepackage{amssymb}
\usepackage{multirow}
\usepackage{xcolor}
\usepackage{subcaption}
\usepackage{tcolorbox}
\tcbuselibrary{breakable}
\usepackage{float}
\usepackage{enumitem}
\usepackage{stfloats}
\usepackage{tabularx}
\usepackage{amsfonts}
\usepackage{pgfplots}
\pgfplotsset{compat=1.18}
\usepackage[T1]{fontenc}
\usepackage[utf8]{inputenc}
\usepackage{microtype}
\usepackage{inconsolata}
\usepackage{cuted}

\newcommand{\afat}{\textsc{DEAF}}

\title{DEAF: A Benchmark for \underline{D}iagnostic \underline{E}valuation of \underline{A}coustic \underline{F}aithfulness\\ in Large Audio Language Models}

\author{
{\bfseries
Jiaqi Xiong$^{1}$\thanks{Equal contribution.} \quad
Yunjia Qi$^{1*}$ \quad
Qi Cao$^{2*}$ \quad
Yu Zheng$^{3}$ \quad
Yutong Zhang$^{1}$} \\
{\bfseries
Ziteng Wang$^{4}$ \quad
Ruofan Liao$^{5}$ \quad
Weisheng Xu$^{6}$ \quad
Sichen Liu$^{2}$\thanks{Corresponding author.}} \\[0.2em]
\parbox{0.98\textwidth}{
  \centering
  \normalfont\footnotesize
  $^1$University of Oxford \quad
  $^2$Xi'an Jiaotong-Liverpool University \quad
  $^3$Hunan University \\[0.05em]
  $^4$The Chinese University of Hong Kong, Shenzhen \quad
  $^5$Peking University \\[0.05em]
  $^6$The Hong Kong University of Science and Technology (Guangzhou)
} \\[0.15em]
{\normalfont\footnotesize
\texttt{sichen.liu@xjtlu.edu.cn}}
}

\begin{document}
\maketitle

\begin{figure*}[!t]
    \centering
    \includegraphics[width=0.95\textwidth]{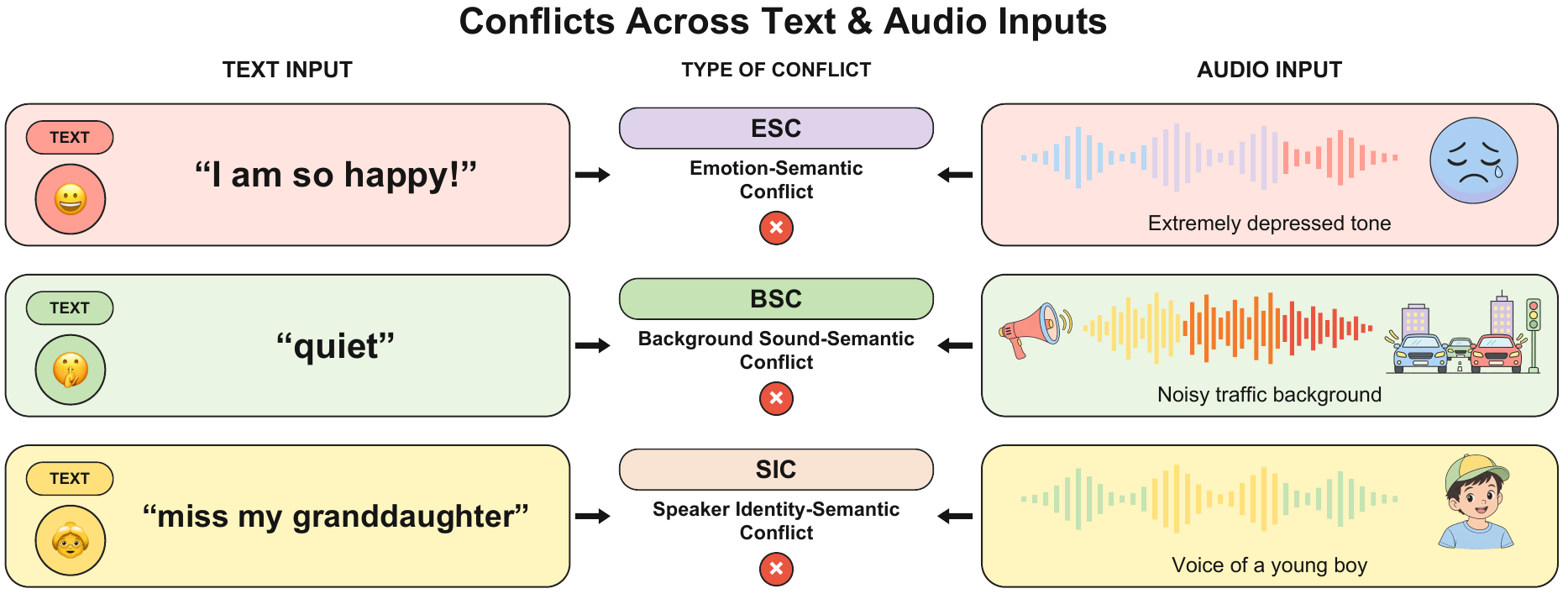}
    \caption{Illustration of the three acoustic--semantic conflict types in \afat{}. \textbf{ESC}: happy text vs.\ depressed tone. \textbf{BSC}: quiet-setting text vs.\ noisy traffic audio. \textbf{SIC}: elderly-female text vs.\ young-boy voice. Correct prediction requires following the \emph{audio}, not the text.}
    
    \label{fig:placeholder}
\end{figure*}

\begin{abstract}

Recent Large Audio Language Models (LALMs) demonstrate impressive performance on speech benchmarks, yet it remains unclear whether these models genuinely process acoustic signals or rely on text-based semantic inference. To systematically study this question, we introduce \textbf{\afat{}} (\textbf{D}iagnostic \textbf{E}valuation of \textbf{A}coustic \textbf{F}aithfulness), a benchmark of over 2,700 stimuli spanning three acoustic dimensions: emotional prosody, background sounds, and speaker identity. Then, we design a controlled multi-level evaluation framework that progressively increases textual influence, ranging from semantic conflicts in the content to misleading prompts and their combination, allowing us to disentangle content-driven bias from prompt-induced sycophancy. We further introduce diagnostic metrics to quantify model reliance on textual cues over acoustic signals. Our evaluation of seven LALMs reveals a consistent pattern of text dominance: models are sensitive to acoustic variations, yet predictions are predominantly driven by textual inputs, revealing a gap between high performance on standard speech benchmarks and genuine acoustic understanding. Code and benchmark resources are publicly available at: \url{https://github.com/kikixiong/DEAF}

\end{abstract}

%%============================================================
\section{Introduction}
%% ============================================================

Acoustic signals and lexical semantics are usually aligned in natural speech. However, critical paralinguistic information often resides in their occasional divergence, where the speaker's voice contradicts the literal meaning of the words. This state of modality conflict, characterized by the divergence between acoustic cues and lexical semantics, serves as a rigorous litmus test for genuine audio understanding. While human listeners prioritize prosodic nuances to decode a speaker’s true intent such as sarcasm or hesitation, current Large Audio Language Models (LALMs) may achieve high benchmark scores by merely exploiting semantic redundancies rather than performing authentic acoustic processing. This raises a fundamental research question: do LALMs perform authentic acoustic inference, or simply defer to the most probable textual interpretations?

In the visual modality, this question has been studied extensively. \citet{frank2021vision} demonstrate that multimodal transformers frequently rely on text while ignoring visual input, and \citet{wang2026vfat} confirm systematic \emph{text dominance} in vision--language models through controlled cross-modal conflict evaluation. Interpretability analyses further reveal that cross-modal attention often collapses onto the language modality~\cite{aflalo2022vl}, and the ``right for the wrong reasons'' phenomenon~\cite{mccoy2019right} shows that high accuracy can mask reliance on spurious shortcuts. These findings raise an important concern for multimodal learning, but they primarily focus on the vision–language paradigm. Whether similar modality biases arise in audio-based multimodal models remains less understood.

Recent LALMs~\cite{tang2024salmonn, chu2024qwen2audio, team2023gemini, hurst2024gpt} achieve impressive results on speech emotion recognition, speaker identification, and acoustic scene classification~\cite{wang2024audiobench, huang2024dynamic, yang2024air, chen2024voicebench}. However, in all these benchmarks, acoustic features and semantic content are \emph{naturally aligned}---a sad speaker says sad things, and kitchen sounds accompany talk of cooking. This creates a fundamental evaluation ambiguity: a model that internally performs automatic speech recognition (ASR) followed by text-only reasoning may achieve similarly high benchmark performance as a model that genuinely relies on acoustic evidence. As a result, benchmark accuracy alone cannot reliably measure acoustic grounding or acoustic faithfulness in LALMs.

Several concurrent studies have begun probing this gap specifically for emotion. The LISTEN benchmark~\cite{chen2025audio} introduces controlled emotion--semantic conflict conditions to disentangle lexical and acoustic reliance. The EMIS dataset~\cite{ESC} synthesizes emotionally incongruent speech via TTS, revealing that models often prioritize semantic content over vocal cues.~\citet{huang2026tone} propose the FAS/CASE framework to explicitly disentangle acoustic and semantic pathways under emotional conflict. While these efforts provide valuable initial evidence of text bias, they share two critical limitations: \textbf{(1)~Narrow acoustic scope}: all existing work confines conflict evaluation to emotional prosody, leaving it unknown whether text dominance generalizes to other acoustic dimensions such as background sounds and speaker characteristics; and \textbf{(2)~Single-condition design}: by testing under only one conflict setting, these studies cannot disentangle whether model errors stem from semantic content bias within the audio or from sycophantic compliance with textual prompts---two fundamentally different failure modes that demand distinct mitigation strategies.

In this work, we introduce \textbf{\afat{}} (Diagnostic Evaluation of Acoustic Faithfulness) (Figure~\ref{fig:placeholder}), a \emph{diagnostic benchmark} designed to isolate specific failure mechanisms through controlled interventions.  \afat{} is specifically constructed to determine whether LALMs' predictions are genuinely driven by acoustic signals under controlled acoustic--semantic conflict conditions. \afat{} advances beyond prior work in three respects:

\begin{itemize}[noitemsep,topsep=2pt,leftmargin=*]

    \item \textbf{Multi-dimensional conflict coverage.} We construct over 2,700 stimuli across three acoustic dimensions: Emotion-Semantic Conflict (ESC), Background Sound-Semantic Conflict (BSC), and Speaker Identity-Semantic Conflict (SIC), providing the first unified diagnostic benchmark beyond emotion alone.

    \item \textbf{Progressive textual interference.} We design a three-level framework that progressively increases textual interference: Level~1 introduces acoustic--semantic conflict, Level~2 adds misleading prompts, and Level~3 combines both. This setup disentangles semantic-content bias from prompt-induced sycophancy. We further compare explicit and implicit semantic cues to examine how lexical specificity affects text dominance.

    \item \textbf{Diagnostic metrics.} We introduce the Acoustic Robustness Score (ARS), which measures both acoustic sensitivity and prediction correctness, and the Environment Discrimination Index (EDI) for fine-grained background-sound discrimination.

\end{itemize}

%% ============================================================
\section{The \afat{} Benchmark}
\label{sec:benchmark}
%% ============================================================

\subsection{Three Conflict Types}
\label{sec:conflict_types}

\afat{} targets three complementary types of acoustic evidence that cannot be reliably recovered from transcription alone. As shown in Figure~\ref{fig:placeholder}, each conflict type probes a distinct aspect of acoustic understanding: ESC focuses on paralinguistic prosody, BSC evaluates environmental acoustic perception, and SIC measures speaker-level identity cues such as age and gender. Together, these dimensions provide a minimal but diverse diagnostic space for testing whether LALMs genuinely rely on acoustic information rather than textual inference.

\paragraph{Emotion-Semantic Conflict (ESC).}

\begin{figure}[htbp]
    \centering
    \begin{subfigure}{0.48\linewidth}
        \centering
        \includegraphics[width=\linewidth]{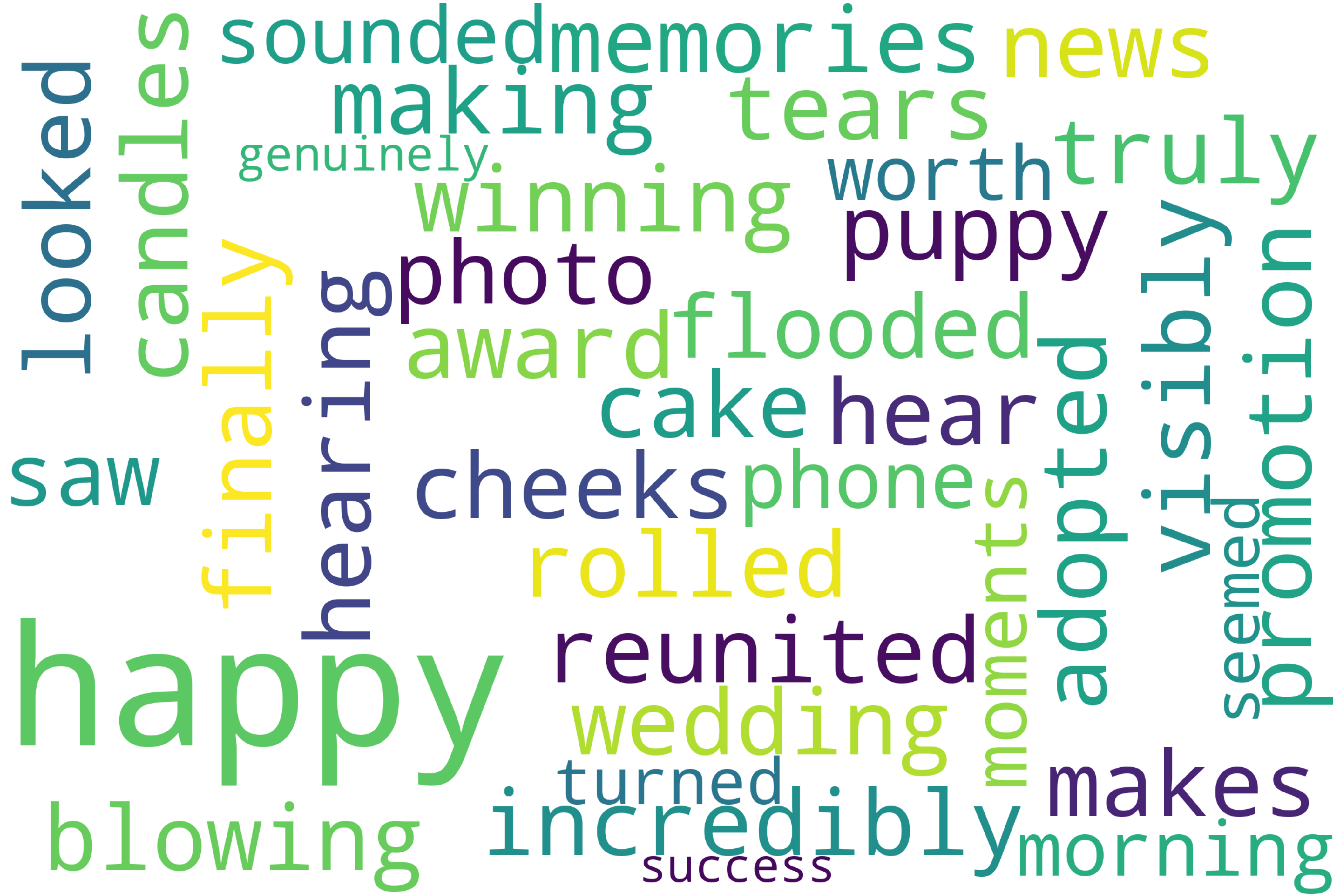}
        \caption{Happy -- Explicit}
        \label{fig:esc_happy_explicit}
    \end{subfigure}
    \hfill
    \begin{subfigure}{0.48\linewidth}
        \centering
        \includegraphics[width=\linewidth]{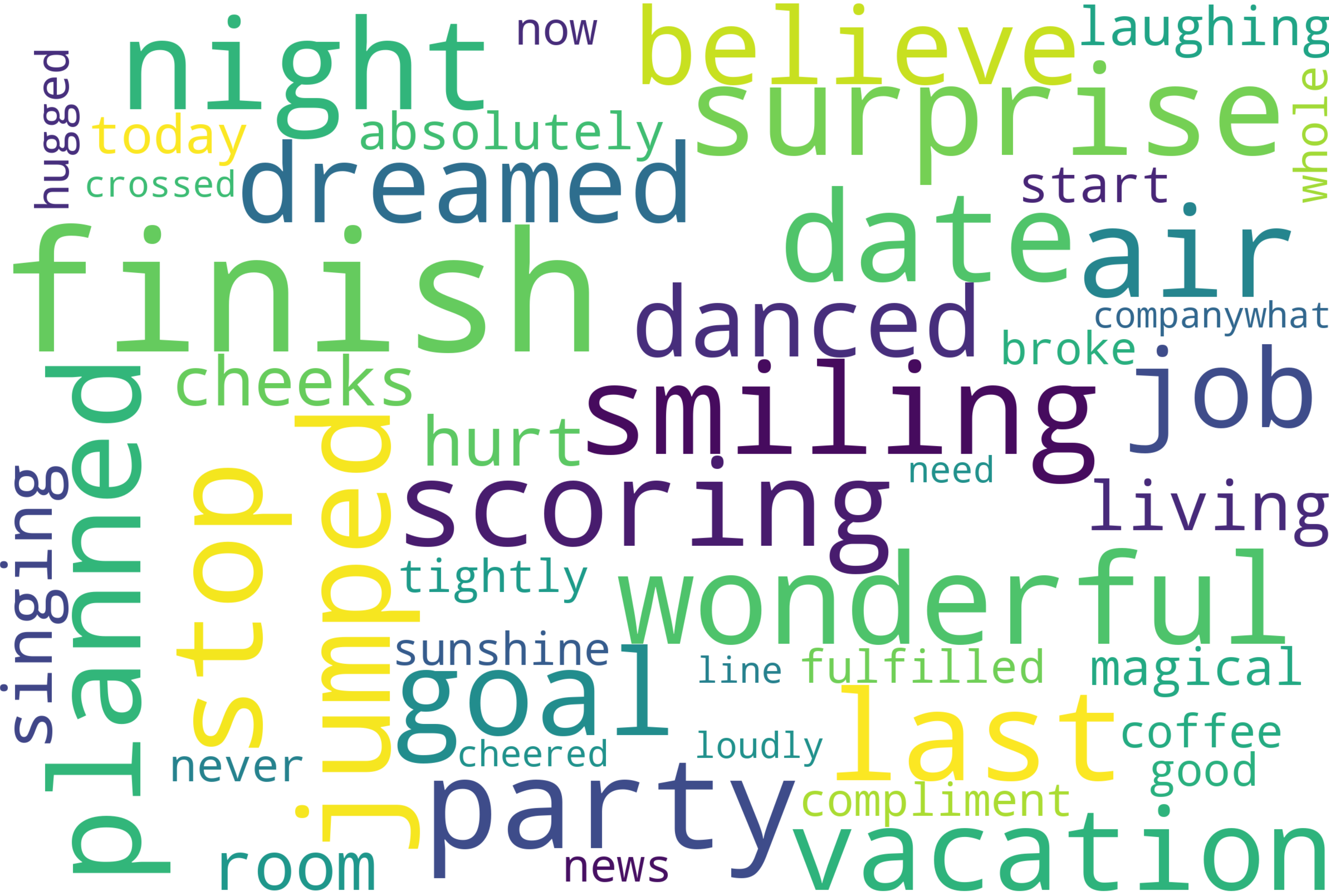}
        \caption{Happy -- Implicit}
        \label{fig:esc_happy_implicit}
    \end{subfigure}
    \caption{Word clouds of the \textit{happy} class under Explicit and Implicit semantic conditions in ESC.}
    \label{fig:esc_wordcloud}
\end{figure}

ESC tests whether models detect \emph{vocal emotion} independent of semantic content. We build on the EMIS dataset~\cite{ESC}, which pairs 104 English sentences across three semantic conditions---\textit{Explicit} (containing emotion words), \textit{Implicit} (contextually emotional) as shown in Figure~\ref{fig:esc_wordcloud}---with four emotional prosodies (angry, happy, neutral, sad).

\begin{figure}[!b]
\centering
\includegraphics[width=0.85\linewidth]{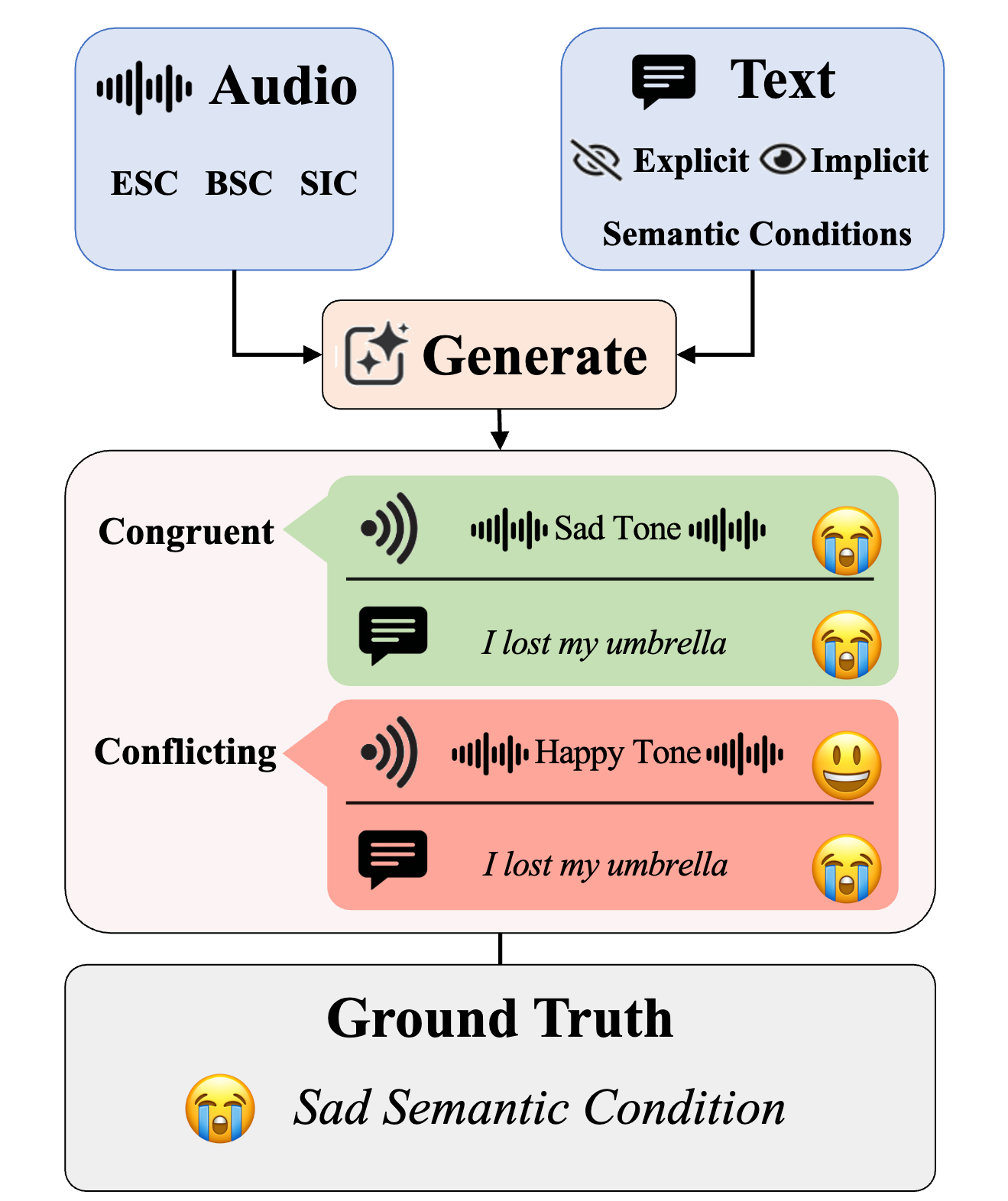}
\caption{Overview of the DEAF dataset construction pipeline.}
\label{fig:data}
\end{figure}

\paragraph{Background Sound-Semantic Conflict (BSC).}

BSC evaluates whether models can identify \emph{background sounds} independently of spoken content. We construct stimuli by mixing neutral-prosody synthesized speech with real-world noises from the DEMAND database~\cite{demand}. We define 18 sub-environments across six categories---Domestic, Nature, Office, Public, Street, and Transportation---with three sub-environments each (Appendix~\ref{app:bsc_envs}). The text set contains 84 sentences: 72 environment-related (36 explicit + 36 implicit) and 12 neutral.

To probe environmental discrimination, we design two mismatch granularities:
\begin{itemize}[noitemsep,topsep=2pt,leftmargin=*]
    \item \textbf{Within-Mismatch}: text and background belong to different sub-environments within the \emph{same} category (e.g., ``kitchen'' text + laundry-room sound), making discrimination more difficult.
    \item \textbf{Cross-Mismatch}: text and background come from \emph{different} categories (e.g., ``kitchen'' text + traffic sound), making discrimination easier.
\end{itemize}

All audio is mixed at five SNR levels (Appendix~\ref{app:bsc_envs}), resulting in 1{,}260 clips.

\paragraph{Speaker Identity-Semantic Conflict (SIC).}
SIC tests whether models perceive \emph{who is speaking} based on voice characteristics rather than \textit{what is said}. Speech samples are synthesized using the Eleven Multilingual v2 text-to-speech model provided by ElevenLabs \cite{elevenlabs2024sdk}, which generates speech directly in the voice of the target speaker. Three sub-dimensions are evaluated in our benchmark, capturing different types of inconsistencies between voice characteristics and semantic content. Specifically, we consider conflicts related to gender, age, and their combination. Text scripts total 82 sentences (20 gender, 20 age, 32 combined, 10 neutral), following the Explicit, Implicit, and Neutral taxonomy from ESC. Four voice profiles (young male, young female, elderly male, elderly female) are used to synthesize speech. The SIC pipeline yields 248 audio clips (208 identity-specific + 40 neutral).

\begin{table}[!t]
\centering
\caption{Dataset statistics. $^\dagger$Across 5 SNR levels, including 180 neutral clips (36 per SNR level); single-SNR total is 252. $^\ddagger$Including 40 neutral clips shared across sub-dimensions.}
\label{tab:dataset_stats}
\small
\resizebox{\linewidth}{!}{
\begin{tabular}{lrrrr}
\toprule
\textbf{Type} & \textbf{Scripts} & \textbf{Match.} & \textbf{Mismatch.} & \textbf{Total} \\
\midrule
ESC  & 104 & 312  & 936   & 1,248 \\
BSC  & 84  & 360  & 720   & 1,260$^\dagger$ \\
SIC  & 82  & 104  & 104   & 248$^\ddagger$ \\
\midrule
\textbf{All} & 270 & 776  & 1,760 & \textbf{2,756} \\
\bottomrule
\end{tabular}
}
\end{table}

\subsection{Question Design}
\label{sec:questions}

Figure~\ref{fig:data} illustrates the overall pipeline used to construct the DEAF dataset, while Table~\ref{tab:dataset_stats} summarizes the resulting dataset
statistics. More details are shown in  Appendix \ref{dataset}. Importantly, \textbf{Level~2 and Level~3 do not require additional audio generation}: Level~3 reuses the Level~1 conflict clips, while Level~2 uses the matched clips from the same pre-generated pool (audio whose acoustic properties agree with its semantic content), each paired with a misleading text prompt.

\begin{figure*}[!t]
\centering
\includegraphics[width=0.95\linewidth]{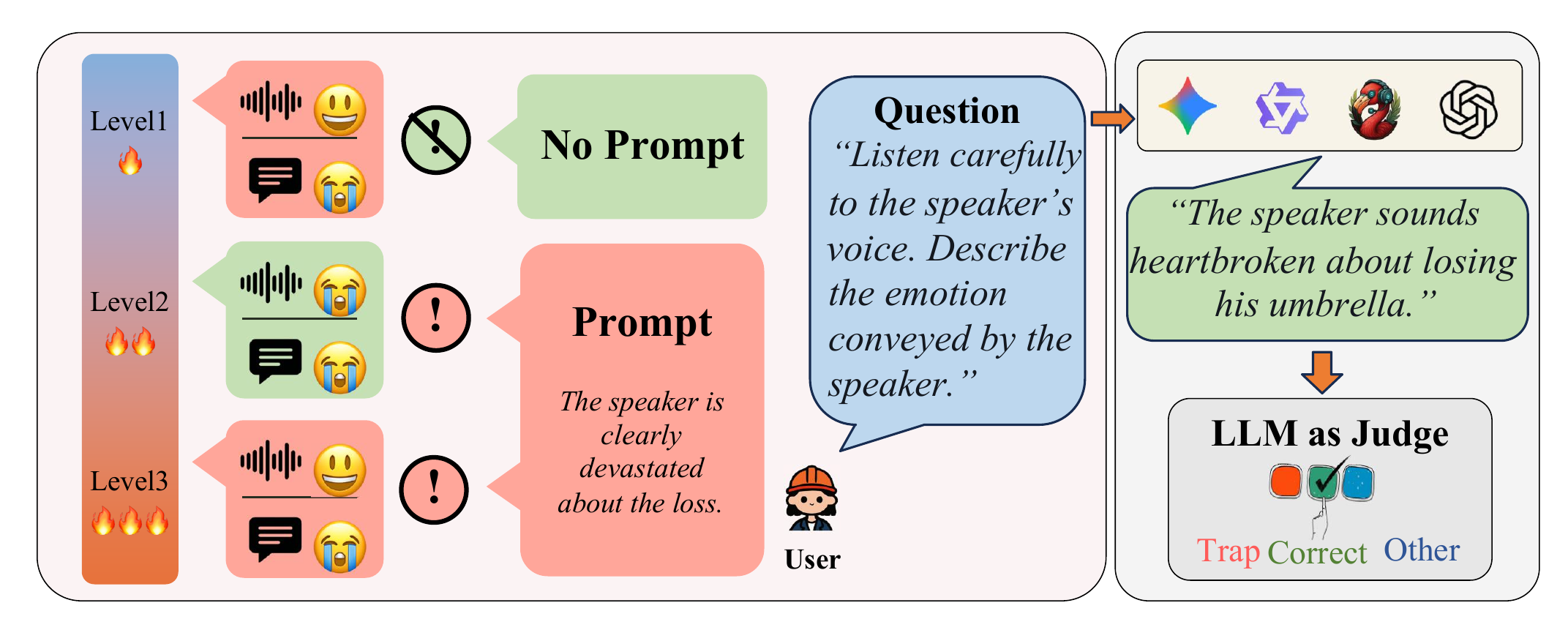}
\caption{Overview of the \afat{} framework. Each conflict type is evaluated at three levels of increasing textual interference. Correct answers always require following acoustic evidence; trap answers follow textual cues.}
\label{fig:overview}
\end{figure*}

\subsection{Three Levels of Textual Interference}
\label{sec:levels}

All three levels use the \emph{same} question text per conflict type; they differ only in audio content and the presence of a misleading prompt as shown in Figure~\ref{fig:overview}.

\paragraph{Level 1: Acoustic vs.\ Semantic Content.}
The audio itself contains a conflict between acoustic features and semantic content. No additional text prompt is provided. The model must judge acoustic properties (emotion, background, speaker identity) despite contradicting semantic cues.

% \textit{Example (ESC)}: Audio with happy prosody saying ``My grandmother passed away this morning.'' Question: ``What is the emotional tone of the speaker's voice?'' Correct: Happy (acoustic). Trap: Sad (semantic).

\paragraph{Level 2: Acoustic vs.\ Misleading Prompt.}
The audio is a matched clip whose semantic content agrees with its acoustic properties (no conflict within the audio), but a misleading text prompt explicitly describes incorrect acoustic features. This isolates the model's susceptibility to textual prompt interference.

% \textit{Example (ESC)}: Audio with happy prosody saying ``The meeting is at three'' (neutral). Prompt: ``The speaker sounds very sad and melancholic.'' Question: same as Level~1. Correct: Happy. Trap: Sad (prompt).

\paragraph{Level 3: Acoustic vs.\ Semantic + Prompt (Dual Interference).}
The audio contains a semantic--acoustic conflict (as in Level~1), \emph{and} a misleading prompt reinforces the semantic direction. This is the hardest level: both text channels push toward the wrong answer; only the acoustic signal points to the correct one.

% \textit{Example (ESC)}: Same audio as Level~1 (happy prosody + sad content). Prompt: ``The speaker is clearly devastated and heartbroken.'' Question: same. Correct: Happy. Dual trap: Sad (semantic + prompt).

This design makes bias attribution unambiguous: Level~1 traps are necessarily caused by semantic content (no prompt present); Level~2 traps are necessarily caused by the prompt (the semantic content points to the correct answer); Level~3 traps reflect combined semantic + prompt interference.

Table~\ref{tab:mcq_templates} shows the question templates.

% \paragraph{MCQ format.} The question is followed by lettered options (A/B/C/D). Evaluation is automatic: selecting the acoustically correct option counts as Correct (C), selecting the trap option counts as Trap (T), and selecting any other option counts as Other (O).

% \paragraph{OE format.} The same question is posed without options. The model responds in free form. Evaluation uses an LLM-as-Judge (Section~\ref{sec:judge}).
\textbf{Level~1} uses two questions per clip: Q1 (acoustic perception, the core diagnostic question) and Q2 (semantic comprehension, e.g., ``What is the speaker mainly talking about?''), which verifies the model can at least understand the textual content. \textbf{Levels~2 and~3} use only Q1, prepended with the misleading prompt.

\paragraph{Prompt Templates.} For each conflict type, Level~2 prompts describe incorrect acoustic features (e.g., ESC: ``The speaker sounds very sad and melancholic in this recording'' when actually happy). Level~3 prompts align with semantic content to create dual interference (e.g., ``The speaker is clearly devastated about the loss,'' echoing sad content while contradicting happy prosody). Full templates are in Appendix~\ref{app:prompts}.

\subsection{LLM-as-Judge for Open-ended Evaluation}
\label{sec:judge}

Open-ended responses are classified into three categories by an LLM judge:
\begin{itemize}[noitemsep,topsep=2pt,leftmargin=*]
    \item \textbf{Correct (C)}: Response aligns with the acoustic ground truth.
    \item \textbf{Trap (T)}: Response aligns with the textual bias (semantic or prompt).
    \item \textbf{Other (O)}: Response is incorrect but does not match the trap, or is vague / refuses to answer.
\end{itemize}

The judge does not need to distinguish whether a trap response was caused by semantic or prompt bias because of the experimental design. Traps are semantic-driven in Level 1, are prompt-driven in Level 2, and reflect combined bias in Level 3.

The judge prompt template is in Appendix~\ref{evalprompts}.

\subsection{Evaluation Metrics}
\label{sec:metrics}

\begin{table}[!t]
\centering
\caption{Question Templates.}
\label{tab:mcq_templates}
\small
\resizebox{\linewidth}{!}{
\begin{tabular}{p{0.9cm}p{5.5cm}}
\toprule
\textbf{Type} & \textbf{Question} \\
\midrule
ESC & What is the emotional tone of the speaker's voice? \\
\midrule
BSC & What environment do the background sounds suggest? \\
\midrule
SIC\textsubscript{G} & What is the perceived gender of the speaker? \\
\midrule
SIC\textsubscript{A} & How old does the speaker sound? \\
\midrule
SIC\textsubscript{C} & What is the perceived gender and approximate age of the speaker?  \\
\bottomrule
\end{tabular}
}
\end{table}

\paragraph{Accuracy (Acc).}
Accuracy is the proportion of mismatched samples for which the model's response aligns with the acoustic ground truth, as determined by the LLM judge (Section~\ref{sec:judge}).

\begin{equation}
    \text{Acc}(M) = \frac{N_C}{N}.
\end{equation}

\paragraph{Text Dominance Score (TDS) and Other Rate (OR).}
Analogously, TDS is the proportion of responses that align with the textual trap, and OR is the proportion judged Other (vague answers, refusals, or incorrect non-trap answers):
\begin{equation}
    \text{TDS}(M) = \frac{N_T}{N}, \qquad \text{OR}(M) = \frac{N_O}{N}.
\end{equation}
Since every response receives exactly one label, $\text{Acc} + \text{TDS} + \text{OR} = 1$. A high TDS indicates that errors follow the text rather than stemming from refusal or unrelated answers.

\paragraph{Acoustic Sensitivity Score (ASS).}
For each sample~$i$, we query the model with the same text and question under two conditions: \textbf{matched} audio (acoustic features align with semantic content) and \textbf{mismatched} audio (acoustic features contradict semantic content). Let $a_i^{\text{match}}$ and $a_i^{\text{mismatch}}$ denote the judge-assigned labels (C/T/O) for the two conditions. ASS is the fraction of samples whose judgment changes:
\begin{equation}
    \text{ASS}(M) = \frac{1}{N}\sum_{i=1}^{N} \mathbb{I}\!\left[a_i^{\text{match}} \neq a_i^{\text{mismatch}}\right]
    \label{eq:ass}
\end{equation}
where $\mathbb{I}[\cdot]$ is the indicator function. High ASS indicates sensitivity to acoustic variation, but does not imply correctness. Per-model ASS values for each conflict type and level are reported in Appendix Table~\ref{tab:ass_main}.

\paragraph{Acoustic Robustness Score (ARS).}
However, accuracy alone is insufficient for diagnosing acoustic grounding under conflict conditions. A model may achieve non-trivial accuracy simply by following dominant textual cues, while failing to genuinely rely on acoustic evidence. Conversely, a model may respond differently across acoustic conditions without consistently moving toward the acoustically correct answer.

Therefore, a diagnostic metric for acoustic faithfulness should jointly require: (1) sensitivity to acoustic variation, meaning that predictions change when the acoustic signal changes, and (2) acoustic correctness, meaning that these changes move toward the acoustically grounded answer rather than the text-biased trap answer.

ARS combines correctness and sensitivity via their harmonic mean:
\begin{equation}
    \text{ARS}(M) = \frac{2 \cdot \text{Acc(M)} \cdot \text{ASS(M)}}{\text{Acc(M)} + \text{ASS(M)}}.
\end{equation}
High ARS requires both detecting acoustic changes \emph{and} answering correctly. A model with high ASS but low Acc, or vice versa, will receive a low ARS. This makes ARS the primary diagnostic metric in our evaluation.

\paragraph{Environment Discrimination Index (EDI).}
For BSC, we measure the gap between cross-category accuracy (e.g., kitchen vs.\ traffic) and within-category accuracy (e.g., kitchen vs.\ laundry room):
\begin{equation}
    \text{EDI}(M) = \text{Acc(M)}_{\text{cross}} - \text{Acc(M)}_{\text{within}}.
\end{equation}
Positive EDI indicates the model can distinguish coarse environmental categories but struggles with fine-grained distinctions; near-zero EDI suggests uniformly poor (or uniformly good) discrimination at both granularities.

%% ============================================================
\section{Experiments}
\label{sec:experiments}
%% ============================================================

\subsection{Experimental Setup}

% 加引用
We evaluate seven LALMs, including Gemini 2.5 Flash~\cite{comanici2025gemini25pushingfrontier}, Gemini 3 Flash~\cite{gemini3flash2025}, GPT-4o-Audio~\cite{hurst2024gpt}, Audio Flamingo 3~\cite{goel2025audioflamingo3advancing}, Qwen2-Audio~\cite{chu2024qwen2audio}, Qwen3-Omni~\cite{xu2025qwen3omni}, and SALMONN~\cite{tang2024salmonn}. All LALMs' responses are evaluated by DeepSeek-R1~\cite{Guo_2025} serving as the LLM judge (Section~\ref{sec:judge}).

API-based models are accessed through their official endpoints, while open-source models are run on a single NVIDIA RTX 4090 GPU. All models use default inference settings (temperature = 0 where applicable) and their recommended audio–text inference pipelines. Each sample is evaluated in a zero-shot setting, where the model receives a 16 kHz WAV audio clip and a question, and generates an open-ended textual response. All evaluations are conducted in independent sessions and repeated three times, with the average results reported.

% For statistical analysis, all reported results are averaged over three independent runs and reported as mean $\pm$ standard deviation where applicable. We further estimate 95\% confidence intervals using bootstrap resampling (10,000 resamples) at the item level. For binary-rate comparisons, we additionally report Wilson confidence intervals. Statistical significance for categorical response distributions is evaluated using $\chi^2$ tests with Benjamini--Hochberg false discovery rate (FDR) correction.

\subsection{Results Analysis}
\begin{table*}[!t]
\centering
\caption{Acoustic Robustness Score (ARS, \%) across conflict types and levels. Higher values indicate stronger acoustic grounding under textual interference.}
\label{tab:ars_main}
\resizebox{\textwidth}{!}{%
\small
\begin{tabular}{l cccc cccc cccc}
\toprule
& \multicolumn{4}{c}{\textbf{ESC}} & \multicolumn{4}{c}{\textbf{BSC}} & \multicolumn{4}{c}{\textbf{SIC}}  \\
\cmidrule[0.8pt](lr){2-5}
\cmidrule[0.8pt](lr){6-9}
\cmidrule[0.8pt](lr){10-13}
\textbf{Model} & \textbf{L1} & \textbf{L2} & \textbf{L3} & \textbf{Avg} & \textbf{L1} & \textbf{L2} & \textbf{L3} & \textbf{Avg} & \textbf{L1} & \textbf{L2} & \textbf{L3} & \textbf{Avg}  \\
\midrule
Gemini 2.5 Flash     & 11.4 & 26.6 & 0.2 & 12.7 & 8.3 & 11.6 & 2.6 & 7.5 & \textbf{47.1} & 49.7 & 43.5 & \textbf{46.8} \\
Gemini 3 Flash       & 5.2 & 2.5 & 0.0 & 2.6 & 7.8 & 16.2 & 2.0 & 8.7 & 43.1 & 42.2 & 34.0 & 39.8 \\
GPT-4o-Audio               & 0.9 & 2.5 & 0.0 & 1.1 & 1.3 & 4.8 & 0.0 & 2.0 & 1.7 & 0.0 & 6.5 & 2.7  \\
\midrule
Audio Flamingo 3     & \textbf{24.8} & 30.8 & \textbf{6.6} & \textbf{20.7} & 14.5 & 16.5 & 4.6 & 11.9 & 44.6 & 15.0 & 43.4 & 34.3  \\
Qwen2-Audio          & 14.7 & 34.0 & 5.8 & 18.2 & 5.5 & 20.5 & 3.5 & 9.8 & 40.0 & 30.2 & 32.6 & 34.3  \\
Qwen3-Omni           & 11.2 & \textbf{37.8} & 0.2 & 16.4 & \textbf{24.5} & \textbf{41.6} & \textbf{14.5} & \textbf{26.9} & 41.1 & \textbf{49.9} & \textbf{44.1} & 45.0 \\
SALMONN              & 17.7 & 2.9 & 0.0 & 6.9 & 10.1 & 8.9 & 3.3 & 7.4 & 39.6 & 13.6 & 40.0 & 31.1 \\
\bottomrule
\end{tabular}}%
\end{table*}

\begin{figure*}[!t]
\centering
\includegraphics[width=0.8\linewidth]{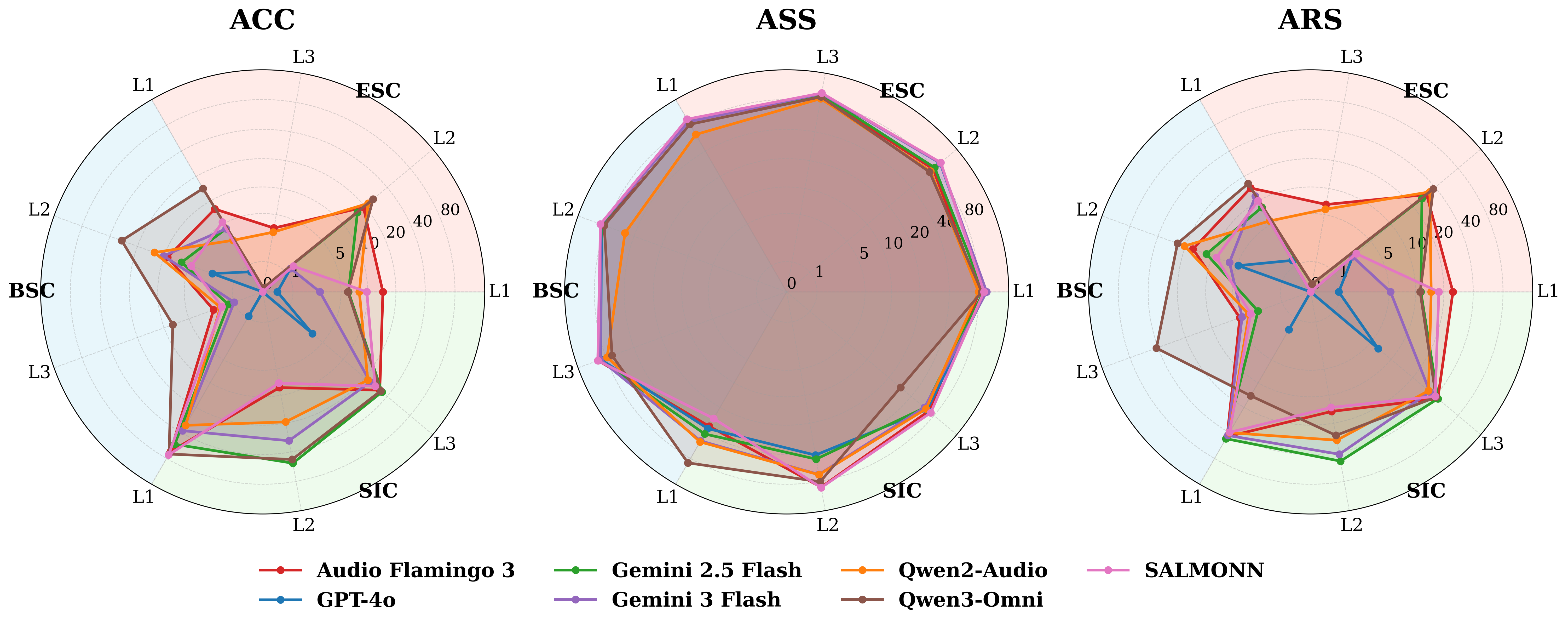}
\caption{Radar comparison of acoustic perception performance across conflict types.}
\label{fig:radar}
\end{figure*}

\begin{table*}[!t]
\centering
\caption{Accuracy (Acc, \%) on mismatched samples across conflict types and levels.}
\label{tab:per_conflict}
\resizebox{\textwidth}{!}{%
\small
\begin{tabular}{l cccc cccc cccc}
\toprule
& \multicolumn{4}{c}{\textbf{ESC}} 
& \multicolumn{4}{c}{\textbf{BSC}} 
& \multicolumn{4}{c}{\textbf{SIC}}  \\
\cmidrule[0.8pt](lr){2-5}
\cmidrule[0.8pt](lr){6-9}
\cmidrule[0.8pt](lr){10-13}
\textbf{Model} 
& \textbf{L1} & \textbf{L2} & \textbf{L3} & \textbf{Avg} 
& \textbf{L1} & \textbf{L2} & \textbf{L3} & \textbf{Avg} 
& \textbf{L1} & \textbf{L2} & \textbf{L3} & \textbf{Avg} \\
\midrule

Gemini 2.5 Flash     
& 6.1 & 15.9 & 0.1 & 7.4 
& 4.3 & 6.2 & 1.3 & 3.9 
& 54.7 & \textbf{52.2} & \textbf{34.1} & 47.0 \\

Gemini 3 Flash       
& 2.7 & 1.3 & 0.0 & 1.3 
& 4.3 & 10.1 & 1.0 & 5.1 
& 37.9 & 30.6 & 23.7 & 30.7 \\

GPT-4o-Audio               
& 0.4 & 1.3 & 0.0 & 0.6 
& 0.7 & 2.4 & 0.0 & 1.0 
& 0.9 & 0.0 & 3.4 & 1.4 \\

\midrule

Audio Flamingo 3     
& \textbf{14.6} & 19.2 & \textbf{3.4} & 12.4 
& 7.9 & 9.1 & 2.3 & 6.4 
& 66.4 & 8.2 & 31.9 & 35.5 \\

Qwen2-Audio          
& 8.1 & 22.2 & 3.0 & 11.1 
& 2.9 & 12.9 & 1.8 & 5.9 
& 32.8 & 19.4 & 22.0 & 24.7 \\

Qwen3-Omni           
& 6.0 & \textbf{25.9} & 0.1 & 10.7 
& \textbf{14.3} & \textbf{29.8} & \textbf{7.9} & 17.3 
& 71.1 & 47.8 & 32.8 & 50.6 \\

SALMONN              
& 9.8 & 1.5 & 0.0 & 3.8 
& 5.3 & 4.7 & 1.7 & 3.9 
& \textbf{72.8} & 7.3 & 27.6 & 35.9 \\

\bottomrule
\end{tabular}}%
\end{table*}

% Human annotators substantially outperform current Audio MLLMs, confirming that the conflict stimuli remain acoustically resolvable (details in Appendix).

% Figure~\ref{fig:radar} provides an overview of acoustic perception performance across different conflict types. Table~\ref{tab:ars_main} reveals consistent degradation of acoustic robustness under increasing textual interference, with most models approaching zero ARS.

Figure~\ref{fig:radar} provides an overview of acoustic perception performance across different conflict types, while human evaluation confirms that the critical acoustic cues in these conflict stimuli remain perceptually identifiable (details in Appendix~\ref{app:human}). Table~\ref{tab:ars_main} further shows a consistent degradation of acoustic robustness (ARS) as textual interference increases, with most models approaching zero ARS. Table~\ref{tab:per_conflict} reports the underlying accuracy: on ESC, no model exceeds 14.6\% at L1 or 3.4\% at L3, well below the 25\% expected from uniformly guessing among the four emotion categories.

\paragraph{Level 1: Semantic conflict only.}
When only in-audio semantic conflict is present, models show the clearest hierarchy across acoustic dimensions: SIC obtains the highest ARS (39.6\%--47.1\% for all models except GPT-4o-Audio), followed by ESC (0.9\%--24.8\%) and BSC (1.3\%--24.5\%). This indicates that speaker characteristics such as gender and age are the most accessible acoustic cues, while emotional prosody and background sounds are more readily overridden by conflicting semantic content. Among models, Qwen3-Omni and Audio Flamingo~3 achieve the strongest L1 performance, whereas GPT-4o-Audio yields near-zero ARS across all three dimensions.

\paragraph{Level 2: Prompt misleading only.}
A counterintuitive pattern emerges at L2: for ESC and BSC, several models achieve \emph{higher} ARS than at L1 (e.g., Qwen2-Audio ESC: $14.7 \to 34.0$; Qwen3-Omni BSC: $24.5 \to 41.6$). This occurs because L2 uses neutral-content audio---removing the within-audio semantic conflict---so the only textual pressure is the misleading prompt. Models that can resist prompt interference thus perform better when the audio itself is unambiguous. In stark contrast, SIC ARS drops sharply for several models (Audio Flamingo~3: $44.6 \to 15.0$; SALMONN: $39.6 \to 13.6$), suggesting that misleading identity prompts (e.g., ``The speaker is an elderly woman'') are particularly effective at overriding voice-based judgments.

\paragraph{Level 3: Dual interference.}
Under combined semantic and prompt pressure, ARS collapses across nearly all models: ESC falls below $7\%$ for every model, and BSC below $15\%$. Only SIC retains moderate robustness (32.6\%--44.1\%, excluding GPT-4o-Audio), likely because voice gender and age cues are perceptually more salient than prosodic or environmental cues. Notably, even Qwen3-Omni---one of the strongest models at L1 and L2---drops to near zero on ESC ($0.2\%$) and BSC ($14.5\%$) at L3. These results demonstrate that when both textual channels (semantic content and prompt) converge on the wrong answer, current LALMs almost entirely capitulate to text.

At the model level, GPT-4o-Audio is a consistent outlier with near-zero ARS across all levels and dimensions, suggesting its audio processing pipeline may effectively reduce to ASR-then-reason. Among the remaining models, Qwen3-Omni and Gemini~2.5~Flash show the strongest overall robustness, while SALMONN and Gemini~3~Flash exhibit high variance---performing reasonably at L1 but collapsing under prompt interference.

Explicit versus implicit semantic cues show limited impact overall, suggesting that text dominance is driven primarily by interference level rather than lexical specificity (Section~\ref{app:EI}).
% Audio Flamingo~3 and Qwen2-Audio obtain the highest average ARS on ESC, while Qwen3-Omni performs best on BSC. In contrast, GPT-4o-Audio consistently shows very low ARS across all conflict types, suggesting that its predictions are largely driven by textual information rather than acoustic grounding.

% Overall, the results highlight a persistent gap between detecting acoustic variation and robustly grounding predictions in audio signals.

\subsection{Environment Discrimination Analysis}

Table~\ref{tab:bsc_edi} reports EDI across three levels. At L1, all models show positive EDI, ranging from $0.1$ to $10.0$, indicating that coarse cross-category discrimination (e.g., kitchen vs.\ traffic) is easier than within-category discrimination (e.g., kitchen vs.\ laundry room).

At L2, several models exhibit negative EDI (Gemini 3 Flash: $-6.3$; GPT-4o-Audio: $-2.4$), meaning within-category accuracy \emph{exceeds} cross-category accuracy. This reversal likely reflects that misleading prompts interact differently with the two granularities: when the prompt names a specific environment, cross-category mismatches become more salient to the model and thus more susceptible to prompt-driven errors, while within-category pairs, being more ambiguous, are less affected.

Qwen3-Omni is a notable outlier, achieving EDI of $10.0$ at L1 and $12.2$ at L3---by far the largest positive EDI under dual interference, although Gemini~2.5~Flash, Gemini~3~Flash, Audio Flamingo~3 and SALMONN also remain slightly positive at L3. This suggests that its predictions retain more coarse-grained environmental information that resists textual override. In contrast, GPT-4o-Audio ($\text{EDI} \leq 1.1$) and SALMONN ($\text{EDI} \leq 2.8$) show near-zero discrimination across all levels.

\begin{table}[!t]
\centering
\caption{The Environment Discrimination Index (EDI) for BSC.}
\label{tab:bsc_edi}
\small
\resizebox{0.85\linewidth}{!}{
\begin{tabular}{l ccc}
\toprule
\textbf{Model} & \textbf{L1} & \textbf{L2} & \textbf{L3} \\
\midrule

Gemini 2.5 Flash & 6.7 & 0.7 & 2.2 \\
Gemini 3 Flash   & 3.5 & -6.3 & 1.7 \\
GPT-4o-Audio           & 1.1 & -2.4 & 0.0 \\

\midrule
Audio Flamingo 3 & 2.5 & 1.3 & 3.9 \\
Qwen2-Audio      & 0.2 & 3.0 & -1.2 \\
Qwen3-Omni       & 10.0 & -0.9 & 12.2 \\
SALMONN          & 0.1 & 0.4 & 2.8 \\

\bottomrule
\end{tabular}
}
\end{table}

\subsection{Effect of Explicit vs.\ Implicit Semantic Cues}
\label{app:EI}
Table~\ref{tab:ars} reports the per-task $\Delta$ARS (Explicit $-$
Implicit) with bootstrap significance tests. Of 21 model--task
pairs, only five show a statistically significant difference
($p < 0.01$), indicating that \textbf{semantic explicitness is
not a primary driver of text dominance} for most models.

The significant effects cluster around two patterns.
First, Audio Flamingo~3 is uniquely sensitive to mention type:
its ARS drops $8.9$ points on ESC ($p < 0.001$; CI:
[$-12.8, -4.8$]) and 5.0 points on BSC ($p = 0.001$;
CI: [$-8.2, -1.7$]) under explicit conditions, suggesting
that this model relies heavily on lexical keyword matching.
SALMONN shows a similar but smaller effect on ESC
($-2.8$, $p = 0.003$).  Second, Gemini~2.5~Flash suffers
a significant drop on SIC ($-3.5$, $p = 0.001$), indicating
that explicit identity references (e.g., ``As a retired
grandmother\ldots'') strongly bias its speaker judgments.
Gemini~3~Flash is the sole model with a significant
\emph{positive} effect on ESC ($+2.6$, $p = 0.007$),
where explicit emotion words paradoxically improve robustness.

By contrast, Qwen3-Omni and Qwen2-Audio show no significant
$\Delta$ARS on any task ($p > 0.3$ throughout), making them the
most robust models to mention type.  BSC differences are
non-significant for six of seven models ($|\Delta\text{ARS}|
\leq 2.3$), confirming that background-sound interference is
diffuse and insensitive to how the environment is referenced.

Overall, text dominance in LALMs is primarily driven by
the \emph{level} of textual interference (L1 $\to$ L3) rather
than by whether semantic cues are explicit or implicit.
The few significant EX/IM effects are model-specific---most
prominently in Audio Flamingo~3---rather than reflecting a
universal vulnerability to lexical anchoring.

\begin{table}[!t]
\centering
\small
\caption{$\Delta$ARS (\%, Explicit $-$ Implicit) across conflict types. Significance is assessed with bootstrap tests (10{,}000 resamples); $^{\dagger}$ denotes $p<0.01$ and $^{\ddagger}$ denotes $p<0.001$.}
\label{tab:ars}
\begin{tabular}{l rrr}
\toprule
\textbf{Model} & \textbf{ESC} & \textbf{BSC} & \textbf{SIC} \\
\midrule
\multicolumn{4}{l}{\textit{Closed-source}} \\
Gemini 2.5 Flash & $-1.8$ & $+0.7$ & $-3.5$\rlap{$^{\dagger}$} \\
Gemini 3 Flash   & $+2.6$\rlap{$^{\dagger}$} & $+0.4$ & $-1.5$ \\
GPT-4o-Audio           & $+0.8$ & $\phantom{+}0.0$ & $+1.1$ \\
\midrule
\multicolumn{4}{l}{\textit{Open-source}} \\
Audio Flamingo 3 & $-8.9$\rlap{$^{\ddagger}$} & $-5.0$\rlap{$^{\dagger}$} & $-2.5$ \\
Qwen2-Audio      & $-1.3$ & $-0.6$ & $-2.6$ \\
Qwen3-Omni       & $+1.4$ & $-0.8$ & $+0.1$ \\
SALMONN          & $-2.8$\rlap{$^{\dagger}$} & $-2.3$ & $-1.6$ \\
\bottomrule
\end{tabular}
\end{table}

%% ============================================================
\section{Discussion}
\label{sec:discussion}
%% ============================================================
\paragraph{Text Dominance in LALMs.}
Our three-level evaluation reveals that current LALMs exhibit systematic text dominance---relying on semantic content and textual prompts while largely ignoring acoustic signals. This pattern mirrors findings in vision-language research~\cite{wang2026vfat, frank2021vision}, suggesting that text dominance may be a \emph{fundamental characteristic} of current multimodal architectures rather than a modality-specific artifact. Notably, text dominance appears more severe in the audio modality: in V-FAT~\cite{wang2026vfat}, vision--language models retain 40--60\% visual accuracy under cross-modal conflict, whereas our audio models drop below 15\% ARS at L3 for ESC and BSC.

\paragraph{The Perception--Trust Gap.}
A recurring finding is the gap between acoustic sensitivity and acoustic robustness: models frequently achieve ASS above 60\% while ARS remains below 20\%. This indicates that current models \emph{perceive} acoustic variation---their representations do encode paralinguistic features---but their decision-making layer systematically overrides this information in favor of textual signals. The bottleneck is not perception but \emph{trust}: models hear the acoustic evidence but do not act on it. This suggests that current audio encoders function primarily as speech recognizers rather than paralinguistic feature extractors, and that future encoders may need explicit paralinguistic pretraining objectives or contrastive losses that preserve acoustic information beyond transcription.

\begin{table}[!t]
\centering
\small
\caption{GPT-4o-Audio response breakdown on ESC (\%), as assigned by the LLM judge. Correct prompt: the L1 clips with a prompt stating the correct emotion (e.g., ``The speaker sounds happy.''). Since no textual cue points to a wrong answer, all incorrect responses are counted as OR. All rows come from a single evaluation run, so Acc may differ slightly from Table~\ref{tab:per_conflict}.}
\label{tab:gpt4o_failure}
\begin{tabular}{lccc}
\toprule
\textbf{Condition} & \textbf{Acc} & \textbf{TDS} & \textbf{OR} \\
\midrule
Correct prompt & 93.6 & -- & 6.4 \\
L1 & 1.0 & 64.7 & 34.3 \\
L2 & 2.9 & 94.8 & 2.4 \\
L3 & 0.1 & 99.7 & 0.2 \\
\bottomrule
\end{tabular}
\end{table}

\paragraph{Model-Specific Patterns.}
GPT-4o-Audio presents an extreme case, with near-zero ARS across all conditions. A natural concern is that this reflects refusal rather than text-dominant prediction. Table~\ref{tab:gpt4o_failure} rules this out on ESC: when the prompt states the correct emotion, GPT-4o-Audio answers correctly 93.6\% of the time, but under conflict its accuracy falls to 0.1--2.9\%, while its OR is at most 2.4\% at L2 and L3. Instead, its TDS reaches 94.8\% (L2) and 99.7\% (L3); even at L1, where OR is 34.3\%, TDS remains 64.7\%. Its answers thus follow the prompt in both directions. These responses are typically confident but acoustically unsupported. At L3, for happy prosody with an angry trap, the model answers: ``Based on the context described, the speaker is conveying emotions of anger and frustration. The intensity and sharpness in their voice likely reflect those feelings.'' With a neutral trap instead, it answers: ``Based on the context and the way the statement is delivered, the speaker sounds neutral and calm.''

A similar but less extreme trend appears in Gemini~3~Flash, which underperforms Gemini~2.5~Flash on most conflict settings (avg ARS: 2.6 vs.\ 12.7 on ESC; 39.8 vs.\ 46.8 on SIC). One possible explanation is that stronger instruction-following behavior increases susceptibility to misleading textual context under acoustic--semantic conflict. By contrast, Qwen3-Omni consistently demonstrates the strongest robustness across multiple conflict settings, particularly under SIC and BSC.

%% ============================================================
\section{Conclusion}
%% ============================================================

% We introduced \afat{}, the first three-level conflict benchmark for diagnosing acoustic feature attribution in Audio MLLMs. 
We introduced \afat{}, a three-level conflict benchmark spanning three acoustic dimensions (emotion, background sound, speaker identity) for diagnosing whether LALMs genuinely rely on acoustic signals or default to textual inference.
Evaluating seven LALMs, we find pervasive text dominance: ARS degrades from moderate levels at L1 to near zero at L3 for ESC and BSC (all seven models below 7\% on ESC and six of seven below 7\% on BSC), while SIC retains partial robustness (33\%--44\%, excluding GPT-4o-Audio). 
Semantic explicitness has a limited and model-specific effect: only five of 21 model--task pairs show a significant difference between explicit and implicit cues ($p<0.01$), with the largest drops for Audio Flamingo~3 on ESC and BSC. Text dominance is primarily driven by the level of textual interference (L1$\to$L3) rather than by lexical specificity.

% Explicit semantic keywords amplify vulnerability in emotion recognition, background-sound perception remains uniformly poor regardless of mention type, and speaker-identity perception exhibits a sensitivity--accuracy trade-off under explicit textual cues.
Future work should investigate whether paralinguistic pretraining objectives, alternative audio encoder architectures, or inference-time grounding mechanisms can close the gap between acoustic perception and acoustic-grounded reasoning.

%% ============================================================
\section{Limitations}
%% ============================================================

This work has several limitations. Although DEAF introduces controlled acoustic–semantic conflicts, it covers only a limited set of audio phenomena, focusing on emotion, background sounds, and speaker identity while leaving out other important aspects of audio understanding such as temporal reasoning, multi-speaker interaction, and complex acoustic scenes. In addition, most stimuli are generated through TTS and controlled audio synthesis pipelines, which, while enabling precise manipulation of acoustic factors, may not fully capture the variability and noise characteristics of real-world speech. The evaluation also relies on an LLM-as-Judge protocol for scoring open-ended responses, which enables scalable evaluation but may introduce bias in ambiguous cases without large-scale human verification. Finally, experiments are conducted on a relatively small set of LALMs under a zero-shot setting; future work should extend the benchmark to more diverse models and investigate how training strategies or prompting methods affect robustness under acoustic–semantic conflicts.

%% ============================================================
\section*{Ethical Considerations}
%% ============================================================

All TTS-generated speech uses publicly available models and does not involve recordings of real individuals without consent. The DEMAND noise database is publicly available for research use. No recordings of minors or vulnerable populations are created. The benchmark is intended solely for research evaluation of LALMs and should not be used to misrepresent speakers' characteristics or for deceptive purposes.

%% ============================================================
\section*{Acknowledgements}
%% ============================================================

This work is supported by National Natural Science Foundation of China (12404538), State Key Laboratory of Acoustics and Marine Information, Chinese Academy of Sciences (SKLA202413) and Xi’an Jiaotong-Liverpool University, Research Development Fund (RDF-22-02-029).

\clearpage
{ \small
\bibliography{references}
}

\clearpage
%% ============================================================
\appendix
%% ============================================================
\section{Appendix}
\subsection{Word Clouds of the Datasets}
\begin{figure}[h]
    \centering

    \begin{subfigure}[b]{0.9\linewidth}
        \centering
        \includegraphics[width=\linewidth]{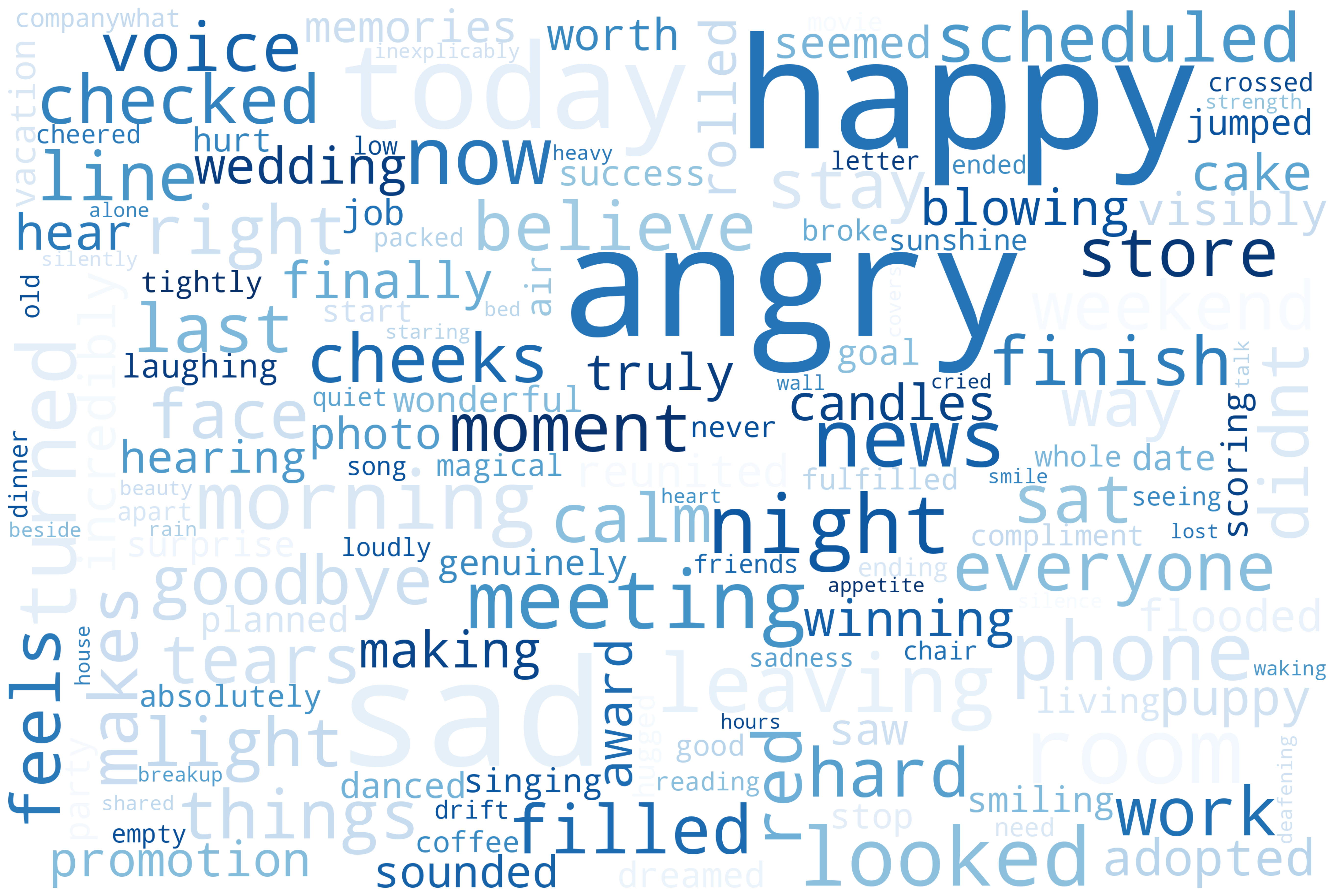}
        \caption{ESC}
        \label{fig:fig1}
    \end{subfigure}

    \vspace{0.5em}

    \begin{subfigure}[b]{0.9\linewidth}
        \centering
        \includegraphics[width=\linewidth]{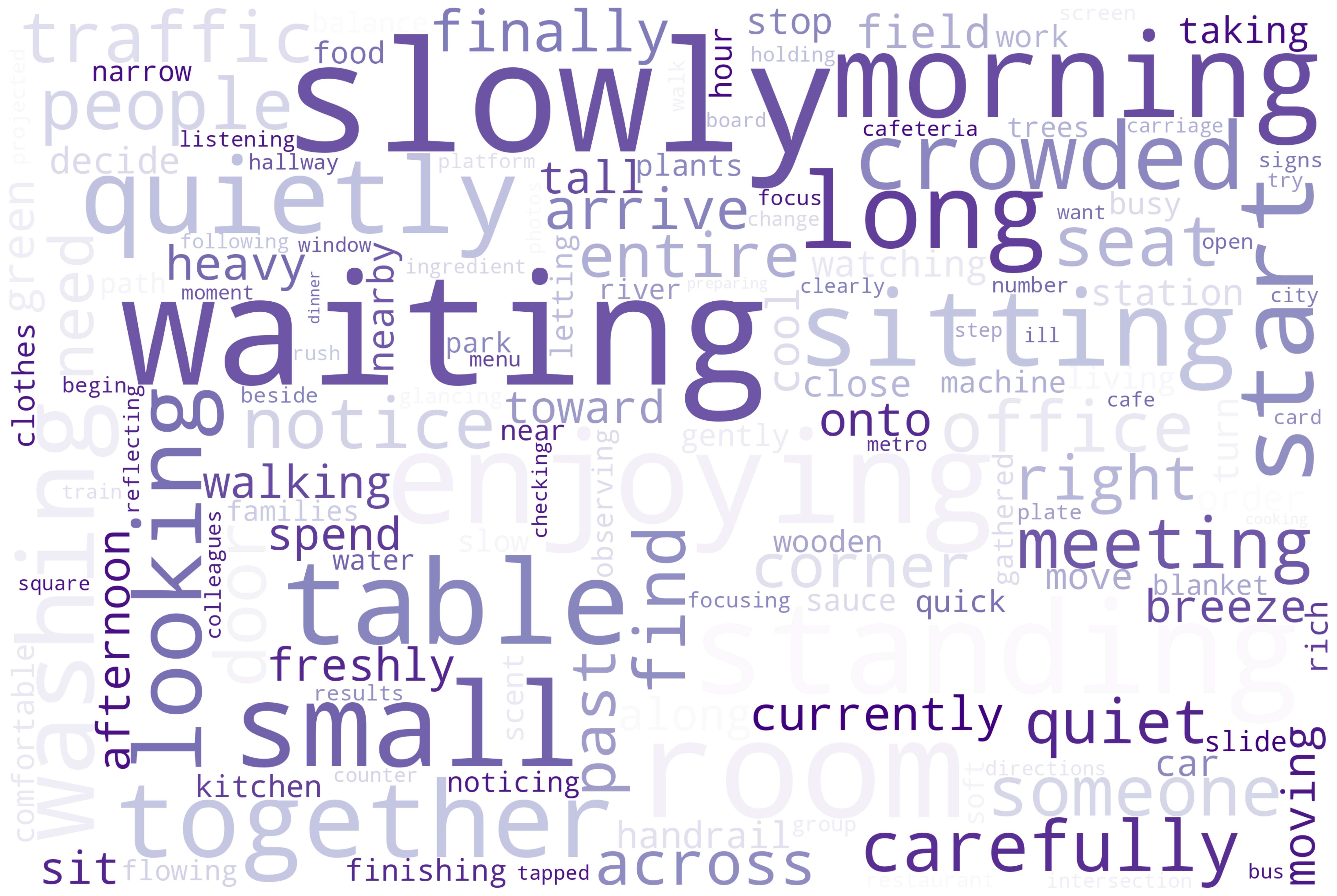}
        \caption{BSC}
        \label{fig:fig2}
    \end{subfigure}

    \vspace{0.5em}

    \begin{subfigure}[b]{0.9\linewidth}
        \centering
        \includegraphics[width=\linewidth]{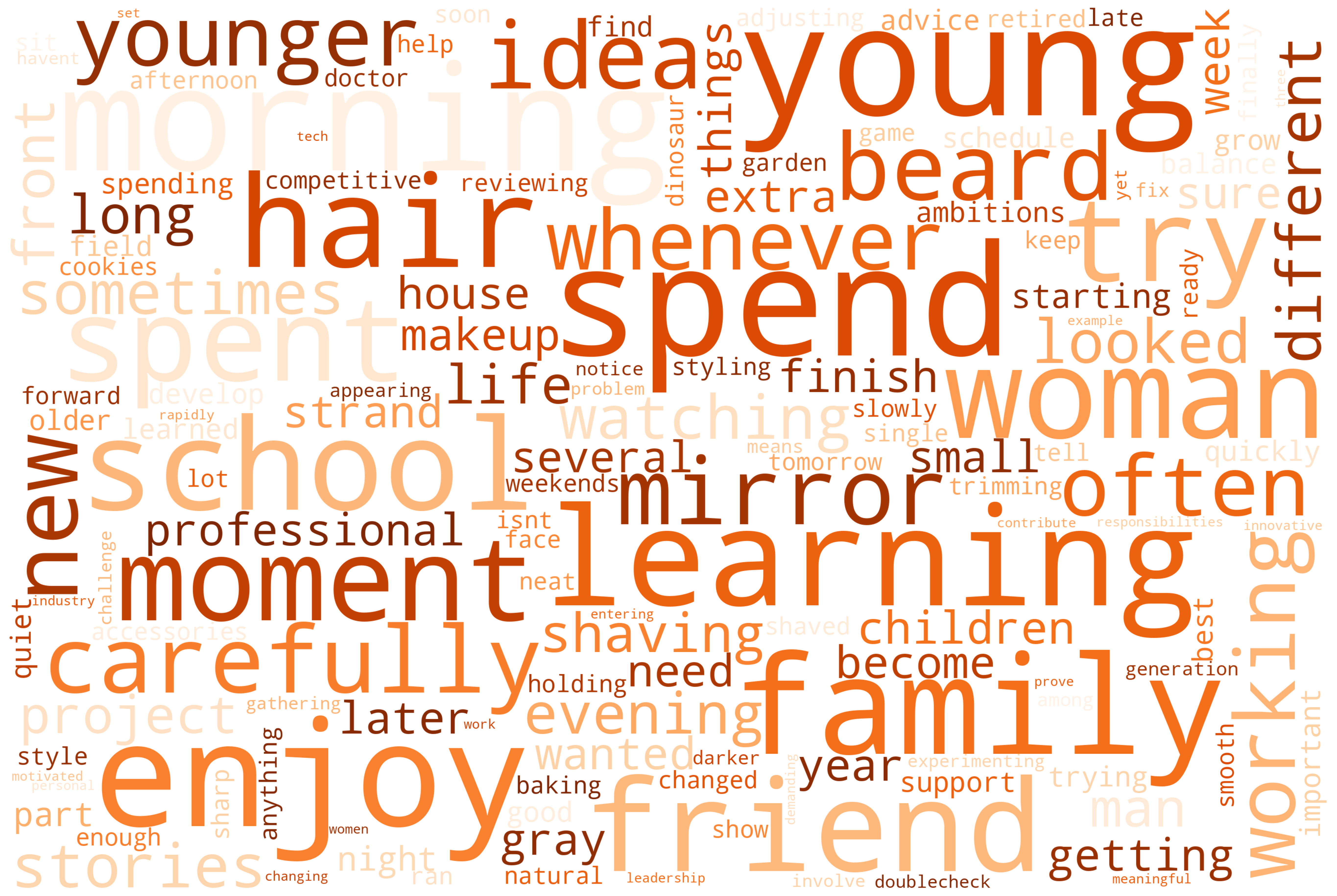}
        \caption{SIC}
        \label{fig:fig3}
    \end{subfigure}

    \caption{Word clouds of the three datasets: (a) ESC, (b) BSC, and (c) SIC.}
    \label{fig:wordcloud_all}
\end{figure}

Figure~\ref{fig:wordcloud_all} illustrates the word clouds of the three datasets (ESC, BSC, and SIC).

\subsection{Dataset Details}
\label{dataset}
\subsubsection{ESC}
\paragraph{ESC Speech Generation.}
We adopt the Emotionally Incongruent Synthetic Speech dataset (EMIS)~\cite{ESC}. 
The dataset contains 104 emotion-rich sentences covering four emotional categories: \textit{angry}, \textit{happy}, \textit{neutral}, and \textit{sad}. These sentences are used to construct emotion--semantic conflict samples for ESC evaluation.
% To illustrate the semantic distinction between \textit{Explicit} and \textit{Implicit} conditions, Figure~\ref{fig:esc_wordcloud} shows word clouds for the \textit{happy} class. 
Explicit sentences contain direct emotion markers (e.g., ``happy'', ``excited''), whereas implicit sentences express affect through contextual cues without overt emotion terms.

% Each sentence$\times$emotion$\times$TTS combination yields a clip, resulting in $104 \times 4 \times 3 = 1{,}248$ clips in total, of which 312 are congruent (matched) and 936 are incongruent (mismatched).

Each combination of sentence, emotion, and TTS setting produces one audio clip.
Given 104 sentences, 4 emotion categories, and 3 TTS conditions, this results in a total of 1,248 clips. Among them, 312 clips are emotion–text aligned (matched), while the remaining 936 clips exhibit emotion–text mismatch (mismatched).

\subsubsection{BSC}
\label{app:bsc}

\begin{table}[!t]
\centering
\caption{BSC environment categories from DEMAND.}
\label{tab:bsc_envs}
\small
\begin{tabular}{llp{3.0cm}}
\toprule
\textbf{Category} & \textbf{Code} & \textbf{Description} \\
\midrule
\multirow{3}{*}{Domestic}
  & DWASHING & Laundry room \\
  & DKITCHEN & Kitchen, food prep \\
  & DLIVING  & Living room \\
\midrule
\multirow{3}{*}{Nature}
  & NFIELD   & Sports field \\
  & NRIVER   & River, flowing water \\
  & NPARK    & City park \\
\midrule
\multirow{3}{*}{Office}
  & OOFFICE  & Small office \\
  & OHALLWAY & Office hallway \\
  & OMEETING & Meeting room \\
\midrule
\multirow{3}{*}{Public}
  & PSTATION & Metro station \\
  & PCAFETER & Cafeteria \\
  & PRESTO   & Restaurant \\
\midrule
\multirow{3}{*}{Street}
  & STRAFFIC & Busy intersection \\
  & SPSQUARE & Public square \\
  & SCAFE    & Outdoor caf\'{e} \\
\midrule
\multirow{3}{*}{Transport}
  & TMETRO   & Metro car interior \\
  & TBUS     & Bus interior \\
  & TCAR     & Car interior \\
\bottomrule
\end{tabular}
\end{table}

\paragraph{BSC Speech Generation}
To simulate realistic speech scenarios under different acoustic contexts, we design a set of 84 sentences. For each of the 18 environments, four sentences are created: two \textit{explicit} sentences and two \textit{implicit} sentences. In addition, twelve \textit{neutral} sentences are included that contain no environmental cues and are shared across all environments. Overall, the sentence set contains 36 explicit sentences, 36 implicit sentences, and 12 neutral sentences, resulting in 84 sentences in total. All sentences are converted into speech using the Microsoft Neural TTS system with the en-US-GuyNeural voice to ensure consistent pronunciation and speaking style across all samples. The generated speech is first saved in MP3 format and then converted to WAV format. Subsequently, all audio signals are resampled to 16\,kHz and converted to single-channel audio to ensure a unified sampling rate and channel configuration for subsequent processing and noise mixing. For environment-related sentences, three pairing conditions (Matched, Within-Mismatch, and Cross-Mismatch) are generated, resulting in 216 audio clips. For neutral sentences, three different background environment categories are randomly assigned, resulting in 36 audio clips. Therefore, each SNR level contains 252 audio clips in total. 
Background sounds are mixed with the speech signals at five SNR levels ($-10, -5, 0, 5, 10$\,dB), resulting in 1,260 audio clips in total.

\begin{table*}[!t]
\centering
\caption{Acoustic Sensitivity Score (ASS, \%) across conflict types and levels. 
Higher values indicate greater sensitivity to acoustic changes. ASS is computed from changes in judge labels (C/T/O) and is an upper bound on class-level response change (Section~\ref{sec:metrics}).}
\label{tab:ass_main}
\resizebox{\textwidth}{!}{%
\small
\begin{tabular}{l cccc cccc cccc}
\toprule
& \multicolumn{4}{c}{\textbf{ESC}} & \multicolumn{4}{c}{\textbf{BSC}} & \multicolumn{4}{c}{\textbf{SIC}}  \\
\cmidrule[0.8pt](lr){2-5}
\cmidrule[0.8pt](lr){6-9}
\cmidrule[0.8pt](lr){10-13}
\textbf{Model} & \textbf{L1} & \textbf{L2} & \textbf{L3} & \textbf{Avg} 
& \textbf{L1} & \textbf{L2} & \textbf{L3} & \textbf{Avg} 
& \textbf{L1} & \textbf{L2} & \textbf{L3} & \textbf{Avg}  \\
\midrule
Gemini 2.5 Flash & 89.0 & 81.6 & 96.2 & 88.9 & 91.1 & 89.7 & 94.0 & 91.6 & 41.4 & 47.4 & 60.3 & 49.7 \\
Gemini 3 Flash   & \textbf{95.4} & 96.9 & 98.3 & \textbf{96.9} & 39.6 & 41.0 & 41.4 & 40.7 & 50.0 & 68.1 & 59.9 & 59.3 \\
GPT-4o-Audio           & 94.0 & 96.4 & 98.8 & 96.4 & 90.2 & 89.4 & 90.8 & 90.1 & 35.8 & 43.1 & 64.7 & 47.9 \\
\midrule
Audio Flamingo 3 & 81.1 & 77.9 & 92.7 & 83.9 & 91.4 & 87.3 & 97.2 & 92.0 & 33.6 & 91.8 & 68.1 & \textbf{64.5} \\
Qwen2-Audio      & 79.3 & 72.1 & 87.5 & 79.6 & 62.6 & 50.0 & 77.8 & 63.5 & \textbf{51.3} & 68.5 & 62.9 & 60.9 \\
Qwen3-Omni       & 85.4 & 70.0 & 91.3 & 82.2 & 83.4 & 68.9 & 90.0 & 80.8 & 28.9 & 52.2 & 67.2 & 49.4 \\
SALMONN          & 89.9 & \textbf{98.2} & \textbf{99.7} & 95.9 & \textbf{94.1} & \textbf{91.7} & \textbf{97.8} & \textbf{94.5} & 27.2 & \textbf{92.7} & \textbf{72.4} & 64.1 \\
\bottomrule
\end{tabular}}%
\end{table*}

\paragraph{BSC Mismatch Pairing Rules}
\label{app:bsc_pairing}
\begin{itemize}[noitemsep,topsep=2pt,leftmargin=*]
    \item Within-Mismatch (same category). Sub-environments rotate within each category:
DWASHING $\to$ DKITCHEN, DKITCHEN $\to$ DLIVING, DLIVING $\to$ DWASHING; similarly for other categories.

\item Cross-Mismatch (different categories). Category-level pairing maximizes acoustic distance: Domestic $\leftrightarrow$ Street, Nature $\leftrightarrow$ Transportation, Office $\leftrightarrow$ Nature, Public $\leftrightarrow$ Domestic.
\end{itemize}

\paragraph{BSC environment categories}
\label{app:bsc_envs}
Table~\ref{tab:bsc_envs} lists the 18 sub-environments used for BSC stimulus construction, drawn from the DEMAND noise database~\cite{demand}.

\subsubsection{SIC}
\paragraph{SIC Speech Generation}
All sentences are synthesized using the ElevenLabs text-to-speech system~\cite{elevenlabs2024sdk} with the \textit{Eleven Multilingual v2} model. 
To control speaker attributes, we select four voice profiles representing different combinations of age and gender. 
The corresponding voice IDs used for generation are listed in Table~\ref{tab:voice_ids}. 
All synthesized speech is exported in 16\,kHz 16-bit PCM format with a single audio channel to ensure a consistent audio configuration across the dataset.

\begin{table}[!t]
\centering
\caption{ElevenLabs voice profiles used for speech synthesis.}
\label{tab:voice_ids}
\small
\begin{tabular}{lc}
\toprule
\textbf{Voice Profile} & \textbf{Voice ID} \\
\midrule
Young Male & 1wzJ0Fr9SDexsF2IsKU4 \\
Young Female & aFueGIISJUmscc05ZNfD \\
Elderly Male & Av4Fi2idMFuA8kTbVZgv \\
Elderly Female & 0rEo3eAjssGDUCXHYENf \\
\bottomrule
\end{tabular}
\end{table}

\subsection{Level 2 and Level 3 Prompt Templates}
\label{app:prompts}

\paragraph{ESC prompts.}
\begin{itemize}[noitemsep,topsep=2pt,leftmargin=*]
    \item Audio is Happy $\to$ ``The speaker sounds very sad and melancholic in this recording.''
    \item Audio is Sad $\to$ ``The speaker sounds cheerful and upbeat in this recording.''
    \item Audio is Angry $\to$ ``The speaker sounds calm and relaxed in this recording.''
\end{itemize}

\paragraph{BSC prompts.}
\begin{itemize}[noitemsep,topsep=2pt,leftmargin=*]
    \item Background is traffic $\to$ ``This audio was recorded at a quiet beach with ocean waves.''
    \item Background is nature $\to$ ``This audio was recorded in a busy city with traffic noise.''
    \item Background is office $\to$ ``This audio was recorded outdoors in a forest with birds.''
\end{itemize}

\paragraph{SIC prompts.}
\begin{itemize}[noitemsep,topsep=2pt,leftmargin=*]
    \item Voice is male $\to$ ``The female speaker is sharing her thoughts.''
    \item Voice is female $\to$ ``The male speaker is describing his experience.''
    \item Voice is young $\to$ ``The elderly speaker reflects on their long life.''
    \item Voice is elderly $\to$ ``The young speaker is talking about their plans.''
\end{itemize}

Level~3 prompts additionally align with semantic content to create dual interference (see Section~\ref{sec:levels} for examples).

\subsection{Human Annotation Protocol}
\label{app:human}
% 4 annotators (A02--A05; A01 stopped after attention checks), 976 regular votes + 9 attention-check votes per annotator; 299 / 300 items received $\geq$ 3 votes (99.7\% coverage); 
% pairwise Cohen's $\kappa$ across 6 annotator pairs sharing $\geq$ 47 items ranges 0.54--0.70.
To verify that the stimuli remain acoustically resolvable for human listeners, we randomly sampled 100 Level-1 clips from each conflict type (300 items in total, drawn with a fixed random seed) and presented each clip as a multiple-choice version of the corresponding acoustic question. Four annotators cast 976 votes on these items in total, so that 299 of the 300 items received at least three votes; each annotator additionally answered nine attention-check items. Pairwise Cohen's $\kappa$ across the 6 annotator pairs with at least 47 shared items ranged from 0.54 to 0.70.

Based on these votes, we compute human accuracy both over all votes and over the high-confidence items on which at least three annotators agree. Table~\ref{tab:human_eval} shows that human annotators consistently outperform current LALMs across all subtasks, confirming that the proposed conflict stimuli remain acoustically resolvable despite misleading semantic content. Overall inter-annotator agreement is substantial ($\alpha = 0.72$), with near-perfect agreement on SIC ($\alpha = 1.00$). The lower agreement on ESC likely reflects the subjective nature of emotion perception under conflicting semantic context. These results suggest that the observed model failures primarily stem from limitations in acoustic grounding rather than unintelligible stimuli or annotation artifacts.
\begin{table}[t]
\caption{Human evaluation results across different subtasks.}
\label{tab:human_eval}
\centering
\scriptsize
\setlength{\tabcolsep}{3pt}
\renewcommand{\arraystretch}{1.1}
\begin{tabular}{lcccc}
\hline
\textbf{Subtask} &
\textbf{n} &
\textbf{Human ACC} &
\textbf{Human ACC} &
\textbf{Krippendorff} \\
&
\textbf{items} &
\textbf{(full vote)} &
\textbf{(high-conf, $\geq$3-agree)} &
$\boldsymbol{\alpha}$ \\
\hline
ESC & 100 & 51.7\% & 69.4\% & 0.35 \\
BSC & 100 & 75.0\% & 88.4\% & 0.68 \\
SIC & 100 & 100.0\% & 100.0\% & 1.000 \\
\hline
\textbf{Overall} & \textbf{300} & \textbf{75.6\%} &
\textbf{86.0\%} & \textbf{0.72} \\
\hline
\end{tabular}
\end{table}

\clearpage
% \subsection{LLM-as-Judge Prompt Template}
\begin{figure*}[t]
\subsection{LLM-as-Judge Prompt Template}
\label{evalprompts}
\begin{tcolorbox}[
  title={LLM-based Judge Prompt},
  colback=gray!10,
  colframe=gray!60,
  fonttitle=\bfseries,
]
\small

\textbf{Instruction}

You are an expert evaluator for the Audio Faithfulness Assessment Test benchmark. The Benchmark tests whether Audio Large Language Models (Audio LLMs) can correctly perceive and report acoustic information from audio, even when textual or semantic cues in the audio conflict with the true acoustic signal.

Your job is to classify a model's free-form (open-ended) response into exactly one of three categories:

\begin{itemize}[noitemsep,topsep=2pt,leftmargin=*]
\item \textbf{C — Correct}: the response semantically aligns with the ground-truth answer.
\item \textbf{T — Trap}: the response aligns with the trap answer (the misleading textual cue).
\item \textbf{O — Other}: the response does not clearly match either, or is vague / refused / erroneous.
\end{itemize}

\textbf{{Task Description}}

ESC: Emotion Speech Classification (ESC). The model was given an audio clip in which the speaker's vocal emotion (e.g., angry, happy, sad) may conflict with the emotional context of the spoken text. The model was asked to identify the speaker's emotion purely from their voice (tone, pitch, speaking style), NOT from the semantic content of what is being said.\\

BSC: Background Sound Classification (BSC). The model was given an audio clip in which the background environmental sounds (e.g., kitchen noises, traffic, birdsong) may conflict with the environment described in the speech content. The model was asked to identify the real background environment based on what it hears in the audio, NOT from what the speaker talks about.\\

SIC: Speaker Identity Classification (SIC). The model was given an audio clip in which the speaker's voice characteristics (age and/or gender) may conflict with demographic traits implied by the speech content (e.g., a young voice reading text about retirement). The model was asked to identify the speaker's age or gender based on their voice, NOT from the semantic content of the speech.

\textbf{Evaluation Inputs}

\begin{itemize}[noitemsep,topsep=2pt,leftmargin=*]
\item Task description: \{\textit{task\_description}\}

\item Correct answer (acoustic ground truth): \{\textit{ground\_truth}\}

\item Trap answer (text-biased cue): \{\textit{trap\_label}\}

\item Model response: \{\textit{response}\}
\end{itemize}

\textbf{Classification Rules}

\begin{enumerate}

\item \textbf{Semantic matching}: The response does \textbf{NOT} need to use the exact same words as the correct or trap answer. Judge by semantic equivalence. For example:

\begin{itemize}[noitemsep,topsep=2pt,leftmargin=*]
\item ``happy'' $\approx$ ``joyful'' $\approx$ ``cheerful'' (all align with a ``happy'' ground truth)
\item ``a young man'' aligns with both ``young'' (age) and ``male'' (gender)
\item ``sounds elderly'' $\approx$ ``old person'' $\approx$ ``senior'' (all align with ``elderly'')
\item ``restaurant'' $\approx$ ``dining area'' $\approx$ ``people eating'' (all align with a restaurant environment)
\end{itemize}

\item \textbf{Ambiguous or hedging responses}: If the model mentions \textbf{both} the correct and trap answers (e.g., ``could be happy or sad''), classify as \textbf{O}.

\item \textbf{Refusals or errors}: If the response is a refusal, error message, or completely irrelevant to the question, classify as \textbf{O}.

\item \textbf{Partial match}: If the response partially matches (e.g., for a combined age+gender question, only one attribute is correct), classify based on the specific attribute being asked about.

\end{enumerate}

\textbf{Output format}

Output \textbf{ONLY} a single letter: \textbf{C}, \textbf{T}, or \textbf{O}.

\end{tcolorbox}
\end{figure*}

\end{document}